\documentclass[final,5p,times,twocolumn]{elsarticle}

\graphicspath{{images/}}

\usepackage{amssymb}
\usepackage{amsmath,bm}
\newtheorem{remark}{Remark}
\usepackage{comment}
\usepackage{bm}
\usepackage{subcaption}

\usepackage[capitalise]{cleveref}

\journal{Chaos, Solitons, \& fractals}

\begin{document}

\begin{frontmatter}



\title{Analysis of a new chaotic system, electronic realization and use in navigation of differential drive mobile robot}


\author[1,2]{Christian Nwachioma}
\author[2]{J. Humberto P\'erez-Cruz}
\address[1]{CIDETEC, Instituto Polit\'ecnico Nacional, Mexico City 07700, Mexico}
\address[2]{ESIME-Azc, Instituto Polit\'ecnico Nacional, Mexico City 02250, Mexico}

\begin{abstract}
This paper presents a new chaotic system having four attractors, including two fixed point attractors and two symmetrical chaotic strange attractors. Dynamical properties of the system, viz. sensitive dependence on initial conditions, Lyapunov spectrum, strangeness measure, attraction basin, including the class and size of it, existence of strange attractor, bifurcation analysis, multistability, electronic circuit design, and hardware implementation, are rigorously treated. Numerical computations are used to compute the basin of attraction and show that the system has a far-reaching composite basin of attraction. Such a basin of attraction is vital for engineering applications. Moreover, a circuit model of the system is realized using analog electronic components. A procedure is detailed for converting the system parameters into corresponding electronic component values such as the circuital resistances while ensuring the dynamic ranges are bounded. Besides, the system is used as the source of control inputs for independent navigation of a differential drive mobile robot, which is subject to the Pfaffian velocity constraint. Due to the properties of sensitivity on initial conditions and topological mixing, the robot's path becomes unpredictable and guaranteed to scan the workspace, respectively.
\end{abstract}


\end{frontmatter}


\section{Introduction}

\label{intro} 
The presence of chaos in a nonlinear dynamical system 
can showcase some unexpected behaviors including having a line of equilibriums \cite{ahmadi2020novel,moysis2019analysis}, multi types of equilibriums \cite{xu20205d} and sensitive dependence on initial conditions. A strict positive element in the Lyapunov spectrum is a necessary but insufficient criterion for chaos \cite{sprott2010elegant}. Furthermore, the system may have unstable fixed points \cite{sambas2020investigation} and still admits bounded aperiodic dynamics due to sequentially counteracting actions including stretching and folding of the dynamics and a free path of flow. Besides, it may as well have no fixed point \cite{zhang2020simple} or have stable fixed points \cite{wang2017chaotic,pham2017generating}, yet it can have a (hidden) chaotic strange attractor. Bounded dynamics and the existence of a strange attractor are consequences of a few other properties of the system, namely topological mixing, and dense periodic orbits \cite{mendez2017new,pham2017different}. Hence, each point on a trajectory of the system eventually comes arbitrarily close to every other point. These features can allow the use of the dynamics of a chaotic system to tactically drive the wheels of a robot to independently cover a workspace. These inherent features can also prove useful in other areas such as secure communications \cite{kondrashov2019application,ccavucsouglu2019new,busawon2018brief,kocamaz2018secure,wang2018novel,bai2018chaos}, robotics \cite{vaidyanathan2017new,zang2016applications}, random number generation \cite{sprott2020chaotic,liao2019design,natiq2019dynamics}, neural networks \cite{de2017stable}, home appliances \cite{nomura1995non}, etc.

Although, it is possible to stabilize a chaotic system to some reference point using the principles of control theory \cite{lai2020infinitely,azar2020stabilization,kamdoum2018dynamic}, the dynamics of a system uncontrolled in the above sense, could be tactically used or the system manipulated for desired behaviors. In \cite{li2018constructing}, low-frequency periodic functions were used in place of the offset boosting parameter of the Sprott case M system \cite{sprott1994some} to produce coexisting attractors. The system was realized in hardware using a field-programmable analog array (FPAA). Moreover, complete amplitude control can be realized using two independent configurable analog modules of an FPAA. Hence, despite the characteristic broadband spectrum of chaotic signals, FPAA can be used to realize an attenuator of amplitude conditioning \cite{li2017linear}, making chaotic signals potentially applicable in radar and communication engineering \cite{wang2015target,liu2007principles,sobhy2000chaotic}. Additionally, starting from the  $3$-dimensional diffusionless Lorenz system \cite{van2000diffusionless}, a linear feedback control input to one of the three equations can be used to extend the system to a hyperchaotic one for some appropriate range of parametric values. Manipulating the polarity preserving nonlinear cross terms can result in a piecewise linear system with hyperchaotic behaviors. Hence, the circuit implementation uses the only diode and does not need multipliers or inductors \cite{li2014new}. In the \cite{petavratzis2020chaotic} the uncontrolled dynamics of a lumped chaotic system consisting of the Logistic and May maps were used to realized path planning for robots with discrete motion directions of $4$ and $8$. In the current study, we shall tactically apply the uncontrolled dynamics of a new chaotic system to a robot with differential motions. The new system has the following interesting and useful attributes: 
\begin{enumerate}
	\item It has four attractors including two fixed point attractors and two chaotic strange symmetric attractors.
	\item The combined basins of attraction of the chaotic strange attractors can encompass roughly $95$ percent of a plane of its equilibriums.
	\item As a control parameter crosses zero, a fixed point is annihilated and recreated afterward, leading to a dramatic change in dynamics. Despite an associated momentary change in topology, chaos persists.
	\item\label{hystereis} Hysteresis is present in the dynamics. Thereby heightening the \textit{randomness} of the system. This is a favorable feature for certain applications requiring, for example, random-like sequence generation \cite{kengne2018dynamics}. 
	\item It is easily realizable by electronic simulators due to its wide basin of attraction. Hence, it can be readily deployed to engineering uses, cf.~\cite{nwachioma2019new}\label{n2}.
\end{enumerate}
We shall rigorously analyze the dynamical properties of the system to fully appreciate its attributes. We shall also realize its electronic circuit design and the corresponding hardware version to demonstrate its viability. Finally, we shall realize the navigation of a wheeled robot with control inputs derived from the dynamics of the system. The robot can independently navigate the workspace.

\section{System description\label{secmath}}
The new chaotic system is given by
\begin{equation}
\begin{split}
&\dot{x} = -a_1x+a_2xz+a_3yz\\
&\dot{y} = a_4y-a_5xz\\
&\dot{z} = -a_6z+a_7xy+a_8z^2\label{sys},
\end{split}
\end{equation}
where $(x,y,z)=:\bm{x}$ is the state vector of the system. This system shows chaotic behavior when
\begin{equation}
\begin{split}
\mathbb{A} := &\{a_1 = 1, a_2 = 1, a_3 = 2.3, a_4 = 2, a_5 = 1,\\&a_6 = 6, a_7 = 1, a_8 =-0.25\}.	
\end{split}
\end{equation}
Elements of $\mathbb{A}$ comprises the system's parameters. The system is invariant on $\pi\enspace rad$ rotation about the $z$-axis. Consequently, given that $(x,y,z)$ is a solution of the system, so is $(-x,-y,z)$. 
Moreover, to ensure uniqueness of system \eqref{sys}, given that $a_{12}$ and $a_{21}$ are entries in a $3\times{3}$ matrix corresponding to the linear part of any $3$-dimensional system, we note that the Lorenz-like systems belong to the class $a_{12}a_{21}>0$, the Chen-like systems belong to the class $a_{12}a_{21}<0$ and the L\"{u}-like systems belong to the class $a_{12}a_{21}=0$ \cite{lu2004new,lu2002dynamical}. System \eqref{sys} belongs to the L\"{u}-like class. In another classification scheme \cite{liu2003new}, considering that the linear part of a $3$-dimensional chaotic system can be represented as $A = diag(a,b,c)$, system \eqref{sys} would satisfy $ab+ac+bc<0$. Hence, it is only similar as in the previous sense, but not equivalent, to the L\"{u} system \cite{lu2002new}. Besides, the L\"{u} system has three equilibrium points, the present system has six. In addition, a chaotic system that is similar but not equivalent, to system \eqref{sys} was presented in \cite{zhou2017new}; they have unequal numbers of equilibriums. Therefore, due to the highlighted differences among all the systems, there is no nonsingular transformation by which any of them can be converted to system \eqref{sys}. Hence, the proposed system is unique.

Next, numerical analyses that serve to shed more light on the dynamical attributes of the system would be presented.

\subsection{Numerical solution}
\begin{figure}[htbp]
	\includegraphics[width=\linewidth]{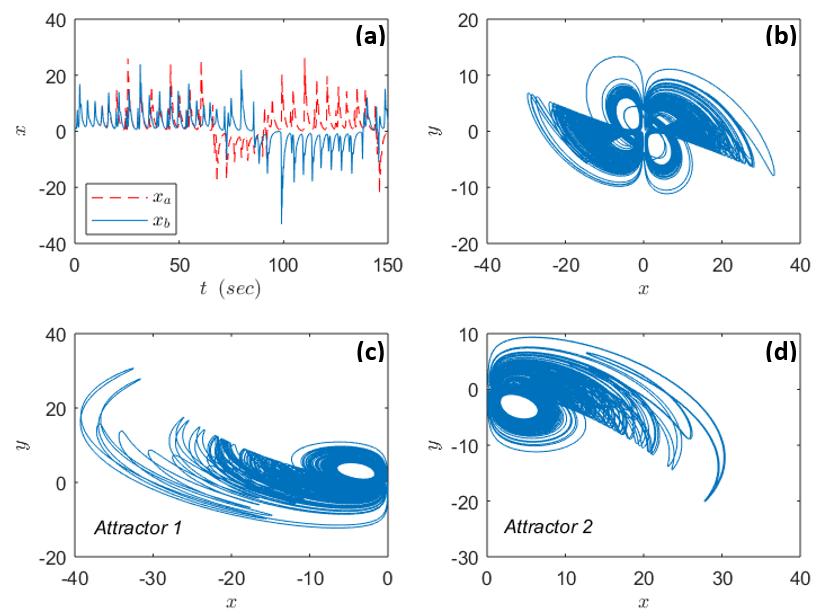}
	\caption{\textbf{(a)} Sensitivity to closely separated initial points $\bm{x}_a(0)=(1,-1,0)$ and $\bm{x}_b(0)=(1,-1,0.001)$ given $a_j\in\mathbb{A}\enspace\forall\enspace j$. \textbf{(b)}: $x$-$y$ phase portrait given $a_j\in\mathbb{A}\enspace\forall\enspace j$ and initial condition is $\bm{x}(0)=(1,-1,0)$. \textbf{(c)}: $x$-$y$ phase portrait given that $a_8 = 1.2$ and initial condition is such that $x\in\mathbb{R}^{-}$, precisely $\bm{x}(0)=(-1,-1,0)$. \textbf{(d)}: $x$-$y$ phase portrait given $a_8=1.2$ and initial condition is such that $x\in\mathbb{R}^{+}$, precisely $\bm{x}(0)=(1,-1,0)$.}
	\label{solution1}		
\end{figure}

\begin{figure}[htbp]
	\includegraphics[width=\linewidth]{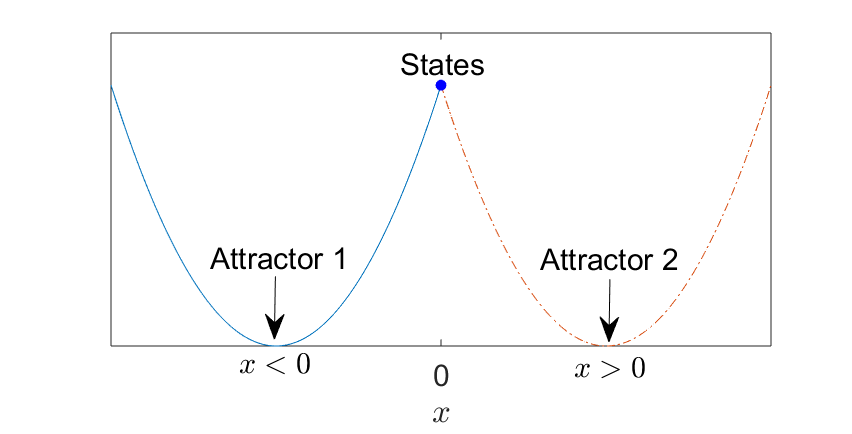}
	\caption{A depiction of bi-stability. If a system is perturbed to the left, it follows \textit{Attractor 1}. If perturbed to the right, it follows \textit{Attractor 2}. Note however, that when one attractor is active, the system can jump to the other if perturbed. That is, it can process a transient, but ultimately identifies one of the attractors.}
	\label{bistability}
\end{figure}

By the method of Runge-Kutta$(4,5)$ pair, a variable-step size of which the maximum step is $0.001$ and a relative tolerance of $2.2204\times10^{-6}$, particular solutions of the system are plotted in \cref{solution1}. \cref{solution1}{(a)} shows sensitive dependence on initial conditions. \cref{solution1}{(b)} is a phase portrait for $a_8<0$. For $a_8>0$: in \cref{solution1}{(c)}, given an initial condition near the attractor such that $x(0)<0$, the attractor has $x(t)<0$ for all $t>0$, in \cref{solution1}{(d)}, given an initial condition such that $x(0)>0$, then $x(t)>0$ for all $t>0$, provided the dynamics is unperturbed. 
That is, the system can trace different attractors depending on the initial condition chosen as depicted in \cref{bistability}. Hence, the system possesses coexisting attractors, with bi-stability appearing when $a_8>0$. This will become more insightful with bifurcation and basin of attraction analyses to be presented later.

\subsection{Analysis of equilibriums}
\begin{figure}[htbp]
	\includegraphics[width=\linewidth]{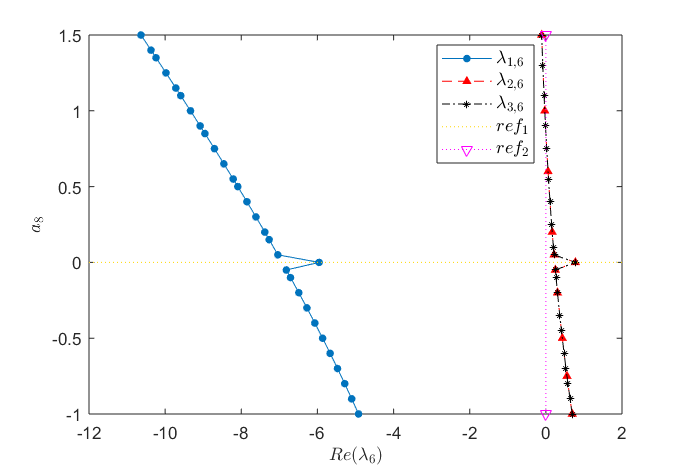}
	\caption{Behavior of $\bm{E}_6$ with varying $a_8$. It shows that $\bm{E}_6$ is a saddle point with two unstable manifolds for $a_8<0.9$ and that it is (marginally) stable for $a_8\ge 0.9$ up to the range of interest. The spike at $a_8=0$ is indicative of a qualitative change in the system's dynamics. This will become clearer with the bifurcation diagram.}
	\label{stability}
\end{figure}

For arbitrary parameters, the equilibrium points of system \eqref{sys} can be analytically obtained by solving the system of equations at points where $\bm{\dot{x}}=0$. Hence, by the \textit{second} equation of the system,
\begin{equation}
\begin{split}
y = {a_5\over a_4}xz\label{x2}.
\end{split}
\end{equation}
Then the \textit{first} equation becomes
\begin{equation}
\begin{split}
0 = -a_1x+a_2xz+a_3{a_5\over a_4}xz^2
\end{split}
\end{equation}
and this gives
\begin{equation}
\begin{split}
x = 0;\quad z = r_{\pm},
\end{split}
\end{equation}
where
\begin{equation}
r_{\pm}=\frac{-a_1a_4\pm\sqrt{a_1^2a_4^2+4a_3a_5a_1a_4}}{2a_3a_5}.
\end{equation}
For $x=0$: by \eqref{x2} and by the \textit{third} equation, now reduced to $-a_6z+a_8z^2=0$, we have 
\begin{equation}
y=0,\quad z=0,\quad z = {a_6\over a_8}.	
\end{equation}		
For $z = r_{\pm}$: using \eqref{x2} and the \textit{third} equation, now $0=-a_6r_{\pm}+a_7x{a_5\over a_4}xr_{\pm}+a_8r_{\pm}^2$, gives
\begin{equation}
x = \pm p_{+}\quad x = \pm p_{-},
\end{equation}		
where
\begin{equation}			
p_{\pm}= \sqrt{\frac{a_6a_4-a_8a_4r_{\pm}}{a_7a_5}}.
\end{equation}	
Using $\pm p_{\pm}$ and $r_{\pm}$ in \eqref{x2} gives 
\begin{equation}
y = \pm q_{\pm}, 	
\end{equation}
where
\begin{equation}
q_{\pm}=\frac{a_5}{a_4}p_{\pm}r_{\pm}. 		
\end{equation}		
The equilibrium points $\bm{E}_k\in\mathbb{R}^3$ are arranged in \cref{equtb}. A Linearized form of the system can be described by the following Jacobian matrix:
\begin{equation}
\begin{split}
J = \left(\begin{array}{ccc} a_{2}\,z-a_{1} & a_{3}\,z & a_{2}\,x+a_{3}\,y\\ -a_{5}\,z & a_{4} & -a_{5}\,x\\ a_{7}\,y & a_{7}\,x & 2\,a_{8}\,z-a_{6} \end{array}\right).
\end{split}
\end{equation}
The Jacobian $J$, can be used to investigate local behavior of the system in the proximity of an equilibrium point. Hence, the system's characteristics equation can be given by
\begin{equation}
det\bigg(I\lambda_k-J(\bm{E}_k)\bigg)=0\label{characteristicseqn},
\end{equation} 
where
\begin{equation}
\lambda_k = \{\lambda_{i,k}\vert i = 1,2,3, k=1,..,6\}\label{eigval}
\end{equation}
is a set of eigenvalues and $I$ is $3\times3$ identity matrix, and using the first Lyapunov method, it can be ascertained that all equilibrium points are unstable given $a_j\in\mathbb{A}$ $\forall$ $j$. However, as $a_8$ is increased within the chaotic space, the symmetric pair $\bm{E}_5$ and $\bm{E}_6$, which is a consequence of the rotation invariance of the system, becomes marginally stable around $a_8\ge0.9$ (see \cref{stability}). Besides, it is important to note that a topological change or bifurcation occurs when $a_8=0$. At this point, a fixed point is annihilated and recreated thereafter but chaos is not lost as a result. However, the dynamics is significantly impacted. This will be more evident in \cref{secbifur} where bifurcation analysis is presented. 
\begin{table}[htbp]
	\centering
	\caption{Equilibriums, $\bm{E}_k (k=1,2,..,6)$, of system \eqref{sys}}	
	\label{equtb}	
	\begin{tabular}{ |l|l|l|l| }				
		\hline
		$ $ & $x $& $y $ & $z$ \\
		\hline
		$\bm{E}_1 $ & $0 $& $ 0$ & $0 $ \\
		\hline
		$\bm{E}_2$ & $ 0$& $0 $ & $a_6/a_8 $\\
		\hline				
		$\bm{E}_3$ & $p_{-} $& $ q_{-}$ & $r_-$\\
		\hline
		$\bm{E}_4$ & $-p_{-} $& $ -q_{-}$ & $r_- $\\
		\hline				
		$\bm{E}_5$ & $-p_{+} $& $ -q_{+}$ & $r_+ $\\
		\hline
		$\bm{E}_6$ & $p_{+} $& $ q_{+}$ & $r_+ $\\
		\hline				
	\end{tabular}
\end{table}
\section{Basin of attraction}
\label{BA}
\begin{figure}[htbp]	
	\includegraphics[width=\linewidth]{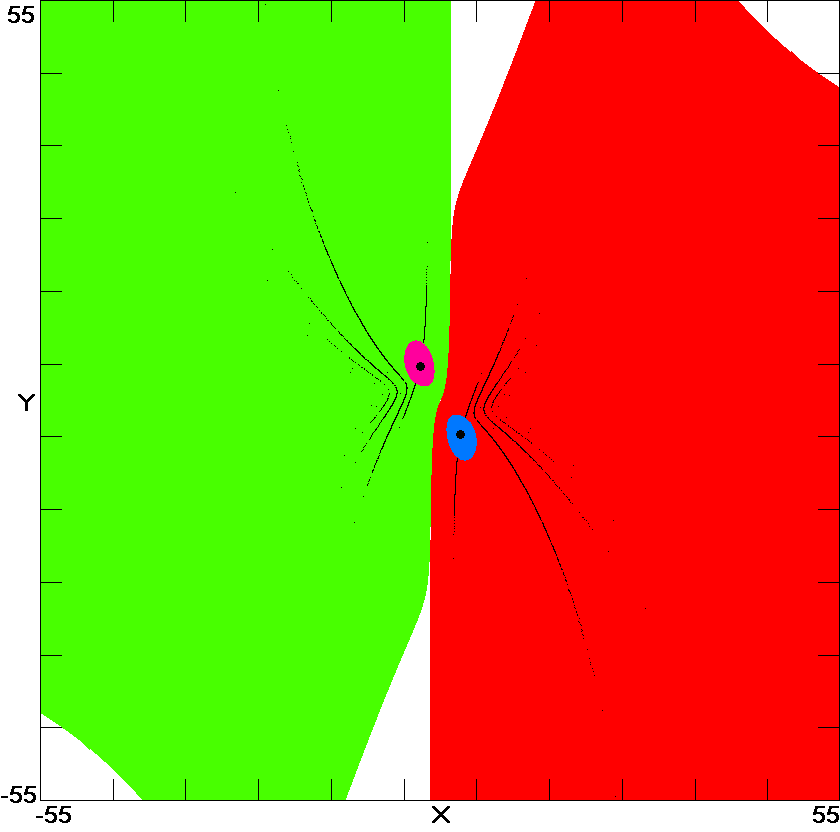}
	\caption{Basins of attraction for the multistable attractors, namely: (1) the green-colored area is the basin of attraction for attractor $1$, initial conditions from this region will identify attractor $1$. The black dot in this region is a fixed point attractor. Initial conditions in the magenta-colored area surrounding the fixed point will converge to the fixed point attractor. (2) the red-colored area is the basin of attraction for attractor $2$, initial conditions in this region will identify attractor $2$. The black dot in this region is a fixed point attractor. Initial conditions in the blue-colored area surrounding the fixed point will converge to the fixed point attractor. Initial conditions in the white (or uncolored) regions will lead to singularities. The black stripes in both regions are the Poincar\'{e} sections. (A web-version or color-print is recommended due to the reference made to colors) -- Figure is courtesy of J. C. Sprott.}
	\label{basinxy}
\end{figure}

Let $\mathcal{A}$ denotes an attractor of system \eqref{sys}. $\mathcal{B} = \{\bm{b}_j\in\mathbb{R}^3\vert j=1,..,m\le N_B\subset\mathbb{R}\}$ is called the basin of attraction given that for any $\bm{x}(0)\in\mathcal{B}$, the steady states of the system are elements of $\mathcal{A}$, $N_B$ is the number of points in the basin of attraction, and $\mathcal{A}\subset\mathcal{B}\subset\mathbb{R}^3$. Also, let us define $\mathcal{B}^{\prime}$ as a set complement of $\mathcal{B}$. Given an $\bm{x}(0)\in\mathcal{B}^{\prime}$, the system will not identify $\mathcal{A}$. Knowledge of $\mathcal{B}$ is important for subsequent calculations. Besides it is useful in engineering applications as it reveals regions of phase space that must be eschewed \cite{sprott2015classifying}. To obtain a pictorial view of a subset of $\mathcal{B}$, the right-hand-side of system \eqref{sys} can be subjected to an \textit{attractor-finding} algorithm. The algorithm deployed is based on the work in ref. \cite{sprott2015classifying}. We look for points in phase space that identify the Poincar\'e section due to each attractor. At the control parameter value of $a_8 = 1.2$, the system has four attractors including two fixed point attractors and two symmetric chaotic attractors. By computing the euclidean distance between non-transient states of the system and the fixed points, we can determine points that lead to the stable fixed points indicated by two black dots in \cref{basinxy}. Also, points that reach the Poincar\'e sections in the plane of the equilibriums $z = -1.4637$, for each chaotic attractor are distinguished, with those identifying attractor $1$, colored green and those identifying attractor $2$, colored red. The white (or uncolored) regions correspond to points that hit singularities or for which the system becomes non-analytic. As the figure shows, points for which $x(0)<0$, do not identify \textit{Attractor 2} and vice versa.%

\begin{figure}[htbp]
	\includegraphics[width=\linewidth]{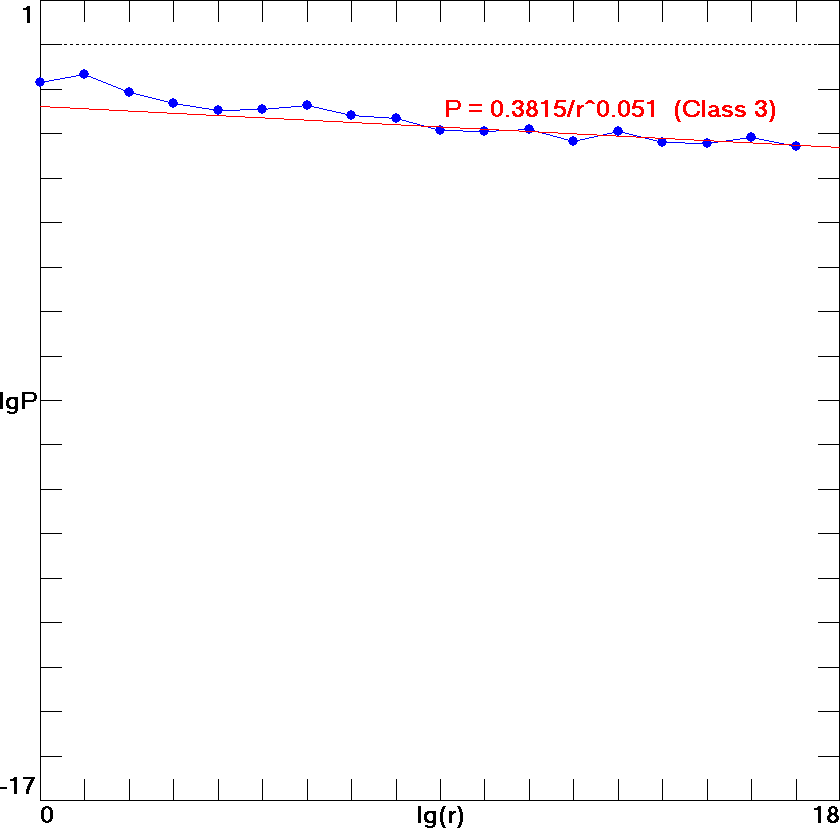}
	\caption{Depiction of basin class and size of the new chaotic system -- Figure is courtesy of J. C. Sprott.}
	\label{basin_size_n_class}
\end{figure}

Moreover, it is equally important to estimate the size of a basin of attraction. Some chaotic attractors have little basins, others have intermediate sizes and stretch to infinity along certain directions, cf.~\cite{perez2020luenberger}. Some others, like the primordial Lorenz chaotic system, have large basins. In fact, the Lorenz system would be globally attracting except for the unstable fixed points at $(0,0,0)$ and $(\pm\sqrt{72},\pm\sqrt{72},27)$ and the uncountably many periodic orbits right in the attractor \cite{sprott2015classifying}. This is particularly important in engineering applications as a large basin reduces or eliminates the danger of electrical perturbations, for example, leading to surges above hardware limits. To classify and quantify a basin of attraction entails calculating the probability $P$ that an initial condition at a distance $r\in\mathbb{R}^{+}$, from the attractor lies within the basin \cite{sprott2015classifying}. At large $r$, the probability follows a power law:
\begin{equation}
P(r) =\frac{P_0}{r^\gamma}
\end{equation}
where $P_0$ and $\gamma$ are the basin quantification and classification parameters. Using these parameters, the basin sizes of chaotic attractors can be grouped into four classes \cite{sprott2015classifying}, viz. 
\begin{enumerate}
	\item Class 1 has $P_0=1$, $\gamma=0$ and attracts almost all initial conditions.
	\item Class 2 has $P_0<0$, $\gamma=0$ and attracts a fraction of the phase space.
	\item Class 3 has $0<\gamma<D$, where $D=3$ in the current case, is the dimension of the phase space and $\gamma$ which is generally a non-integer, is the dimension of the phase space not in the basin.
	\item Class 4 has $\gamma = D$ and has a basin that is bounded and does not stretch to infinity in any direction.
\end{enumerate}
In the computation of the basin class and size of system \eqref{sys}, very many distances $r$, stretching up to $10^{18}$ are computed and their corresponding probabilities calculated. The probability is a measure that a point, which is a distance of $r$ from the attractor, lies in the basin of attraction. To aid visualization of the huge collection of data points, logarithmic scales are used on both axes in \cref{basin_size_n_class}. The figure shows only a few markers. Besides, the scale allows us to easily fit a line from which a slope and an intercept are estimated. These parameters are used to compute $\gamma=0.051$ and $P_0=0.3815$, respectively. Hence, the system has a Class 3 basin. Since the basin stretches to computational infinity and has a codimension of $0.051$, we can make the deduction that the composite basin of attraction of the system takes up to roughly $95$ percent of the current plane, which is good and high enough for most engineering applications.

\section{Chaotic strange attractor}
To address every term in the phrase, \textit{chaotic strange attractor}, we carry out three tasks viz.,
\begin{enumerate}
	\item compute a positive Lyapunov characteristic exponent.
	\item show that the dimensionality of the steady-state solution is fractional.
	\item show that the trace of the Jacobian of the system is negative. 
\end{enumerate}
\subsection{Finite-time Lyapunov spectrum\label{secLyap}}
\begin{figure}[htbp]
	\includegraphics[width=\columnwidth]{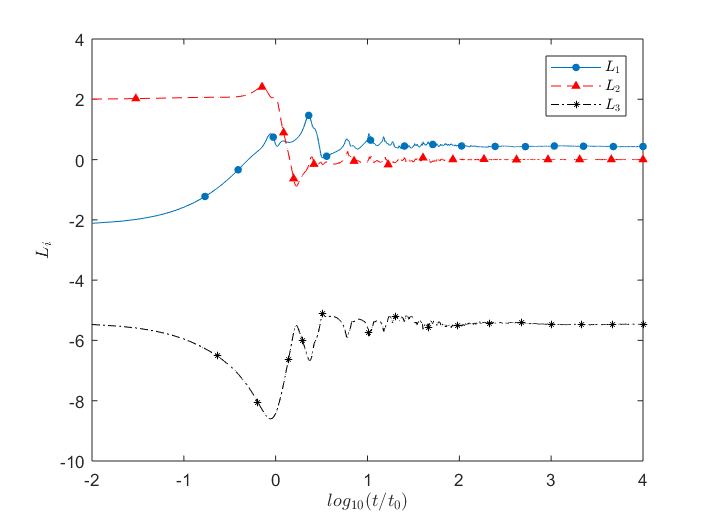}
	\caption{Lyapunov spectrum computed by solving system \eqref{sys} for $a_j\in\mathbb{A}$ and with the iterations starting at $\bm{x}(0)=(0.1001, 0.1003, 0.1003)$. Time starts at $t=0$, and with a step of $0.01$ up to $t=10,000$. Giving a total of 1 million iterations. $t_{0}$ is a unit of time, which serves the purpose of making the horizontal axis dimensionless. Logarithmic scale on the axis is for better visualization.}%
	\label{lyap}
\end{figure}

For an $n$-dimensional continuous autonomous system, the Lyapunov spectrum is a set comprising $n$ Lyapunov characteristic exponents $L_j$, $j=1,2,...,n$. And the Lyapunov characteristic exponent is a measure of the exponential rate of separation of infinitesimally close trajectories in phase space. A positive exponent indicates the possibility of chaos provided solutions of the system are not asymptotically periodic. For $n=3$, the system is chaotic if $L_1>0$, $L_2=0$ and $L_3<0$ with $|L_1|<|L_3|$ \cite{wang2012chaotic}. Thus, the sum of the exponents is negative implying that the system is dissipative. On the other hand, if $|L_1|=|L_3|$, the sum of the exponents is zero, and the system is said to be a Hamiltonian system. The magnitude of the positive exponent is a measure of the rate of expansion along a certain direction in phase space. In another vein, given that two exponents are negative and the other is zero, this is indicative of a limit cycle \cite{muthuswamy2010simplest}. Moving forward, Lyapunov spectra are computed for system \eqref{sys} using several initial conditions, time steps, and various numbers of iterations to ensure these factors would not meaningfully affect the convergence. The displayed Lyapunov spectrum in \cref{lyap} can be obtained using an initial condition on the attractor, e.g.  $\bm{x}(0)=(0.1001, 0.1003, 0.1003)$, $a_j\in\mathbb{A}$ for all $j$ and performing $10^6$ iterations. The Lyapunov exponents are calculated according to ref. \cite{wolf1985determining} and the results are $L_1=0.475$, $L_2 = 0.000$, and $L_3=-5.509$. $L_1$, since it is positive, is indicative of the possibility of chaos. 

\subsection{Lyapunov dimension}
The Kaplan-Yorke formula \cite{frederickson1983liapunov}:
\begin{equation}
D_{KY} = j + \frac{L_1+..+ L_j}{L_{j+1}}
\end{equation}
can be used to calculate the Lyapunov dimension from the finite-time Lyapunov spectrum. $j$ is an integer which is the maximum $m < n$ such that the sequential sum $L_1+...+ L_m\ge0$. For the system at hand, $L_1+L_2 = L_1 >0$ so that $j=2$ and $L_{j+1} = L_3 = -|L_3|$. Hence, we can estimate the dimensionality of the steady state solution using the expression
\begin{equation}
D_{KY} = 2 - \frac{L_1}{|L_3|},
\end{equation}
since $L_2=0$ for the chaotic flow \cite{sprott1994some}. Therefore, the Lyapunov dimension of the steady state solution is $D_{KY}=2.086$. A fractional $D_{KY}$ is a measure of strangeness. It also indicates that the system is not Hamiltonian, in which case $D_{KY}$ would be an integer that is equal to the phase space dimension \cite{IJET16826}.

\subsection{Existence of attractor\label{phasevol}}
The steady state volume of an attractor is zero. To compute the volume of the present system, we can start by investigating the impact of the flow on a volume element in phase space. Let us consider that system \eqref{sys} can be written as
\begin{equation}
\dot{\bm{x}} = \bm{f}(\bm{x}),
\end{equation}
where $\bm{f}(\bm{x})=(\dot{x},\dot{y},\dot{z})^T$. Let the flow of $\bm{f}$ be $\phi_t$ and let the \textit{divergence} of the flow be $\eta$, that is
\begin{equation}
\eta = \nabla\cdot\bm{f}(\bm{x}).  
\end{equation}
Let $D$ be a smooth and connected subset of $\mathbb{R}^3$ and let $D(t) = \phi_t(D)$ be the image of $D$ under the flow. Finally, let $V(t)$ be the volume of $D(t)$, then according to the Liouville's theorem, we can write
\begin{equation}
\dot{V}(t) = \int_{D(t)}\nabla\cdot\bm{f}(\bm{x})dxdydz = \eta V(t)\label{Liouville}.
\end{equation}
Assuming that $\eta$ is a constant, solving the first order ordinary differential equation \eqref{Liouville},  it can be seen that an arbitrary volume will evolve as
\begin{equation}
V(t) = V_0\exp(\eta t)\label{voldynamic}.
\end{equation}
If the system is divergenceless, i.e. $\eta=0$, then $V(t) = V_0$ for all $t>0$ and the system is said to conserve volume. This is the situation for Hamiltonian chaotic systems \cite{IJET16826}. If $\eta>0$, then $V(t)\rightarrow\infty$ as $t\rightarrow\infty$, and the flow is said to expand volume. This can be observed in a system with stretching (and perhaps weak or no folding) action in its dynamics. Finally, if $\eta<0$, then $V(t)\rightarrow 0$ as $t\rightarrow\infty$, and the system is said to be dissipative \cite{yu2019analysis}. A system satisfying the latter, is said to have an attractor.

Unfortunately, we cannot expressly ascertain the impact of system \eqref{sys} on a volume element since
\begin{equation}
\eta =(a_2+2a_8)z-(a_1-a_4+a_6)\label{gradv1}
\end{equation} 
is state-dependent. To avoid this drawback, it can be noted that we can also have \cite{sprott1994some}
\begin{equation}
\eta := tr(J) = \sum_{j=1}^{3}L_j\label{sumLyap}. 
\end{equation} 
From result in \cref{secLyap}, it can be seen that the sum in \eqref{sumLyap} is negative, which implies that the flow generated by the system contracts a volume in phase space going by \eqref{voldynamic}. Therefore, a chaotic attractor, which corresponds to the steady state solution, exists.

Finally, based on results in this section, we can conclude that system \eqref{sys} has a chaotic strange attractor. 

\section{Bifurcation analysis\label{secbifur}}
Bifurcation reveals a qualitative or topological change in the behavior of a dynamical system as the system's (bifurcation) parameter is slowly varied. A bifurcation diagram can reveal the presence of periodic orbit, limit cycle, or chaotic orbit. It provides a visual clue of the system's solutions across a parametric range of interest.  

\begin{figure}[htbp]
	\includegraphics[width=0.9\linewidth]{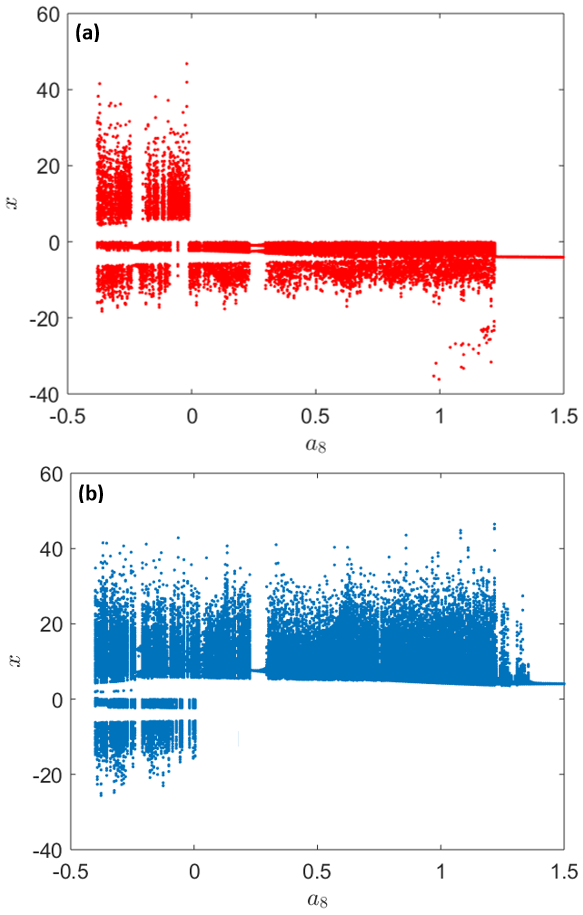}
	\caption{The local maxima of the coordinate $x$ plotted against the varying control parameter, $a_8$. \textbf{(a)}: The chosen initial condition for the first iteration was such that $x(0)<0$. Precisely, $\bm{x}(0)=(-2.1441,-0.3086,0.1113)$ was used. It can be noticed that for this choice and except for possible transient effects, points on the attractor are such that $x(t)<0$ for $a_8>0$. \textbf{(b)}: The chosen initial condition for the first iteration was such that $x(0)>0$. Precisely, $\bm{x}(0)=(2.1441,-0.3086,0)$ was used. It can be noticed that for this choice and with the exception of possible transient effects, points on the attractor are such that $x(t)>0$ for $a_8>0$.}
	\label{bfa8}
\end{figure}

System \eqref{sys} is analyzed with $a_8$ as the control parameter. It is varied in the interval $[-0.5,1.5]$. To visualize the property of hysteresis in the system, we can either perform a forward and backward continuation of the control parameter or we can appropriately choose the initial conditions assuming we know a priori how they impact the system behaviors. Using the latter, first, we choose an initial condition, such that $x(0)<0$, precisely $\bm{x}(0)=(-2.1441,-0.3086,0.1113)$. The system is solved using the Runge-Kutta(4,5) pair and the local maxima are sorted from the states. The final state of each iteration is used as the initial condition for the next. Assuming we are performing forward continuation, at the first instance when $a_8>0$, we can make $x(0)$ negative. However, for backward continuation, this would not be necessary as can be deduced from the bifurcation diagram. In \cref{bfa8}{(a)}  is a trace of the local maxima of $x(t)$ for forward stepping of the control parameter, cf.~\cite{lai2016research}. Where $a_8>0$, it can be seen that starting at any initial conditions for which $x(0)<0$ and using the final state of an iteration cycle as the next initial condition, all future local maxima of $x(t)$ are on the lower half-plane. An analogous situation can be observed in \cref{bfa8}{(b)} in which the iteration was started at $\bm{x}(0)=(2.1441,-0.3086,0)$. It is important to note that where $a_8<0$, local maxima of $x(t)$ appear on both sides of $\mathbb{R}$, the initial conditions used notwithstanding. Traversing the parameter domain from right to left, at $a_8=0$, a fixed point is annihilated and recreated immediately afterward. Even though chaos has persisted at the occurrence of the topological event, a few dynamical properties have changed. For instance, the bifurcation ends the bi-stable hysteresis and the coexisting attractors coalesce. The present discourse should be insightful with regard to the unique solutions of the system given in \cref{solution1}. In conclusion, it can be seen that except for a few periodic windows, chaos persists for $a_8\in{[-0.38,1.3]}$.

\section{Electronic circuit design and  implementation}
\label{circuitdesign1}
\begin{figure}[h!]
	\includegraphics[width=\linewidth,height=0.7\textheight]{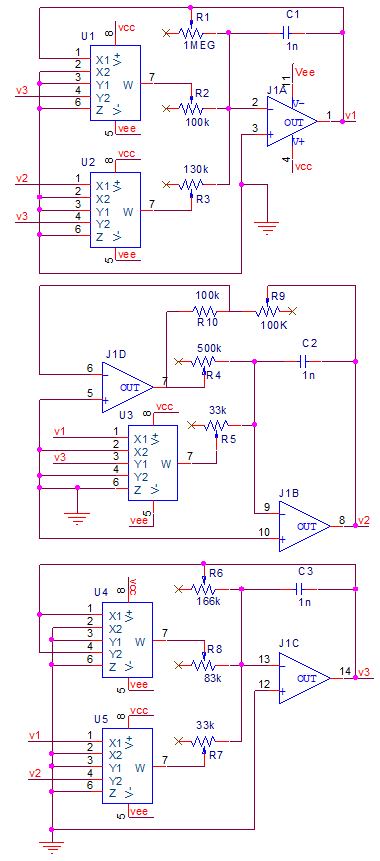}
	\caption{Electronic circuit schematic of system \eqref{sys2}. Currently, $R_8=83k\Omega$. To experiment with $R_8=400K\Omega$, the output of the unit U4 should be inverted. Vcc and Vee are $+15$ and $-15$ \textit{volts}, respectively.}
	\label{ecct}
\end{figure}

\begin{figure}[htbp]
	\includegraphics[width=\linewidth]{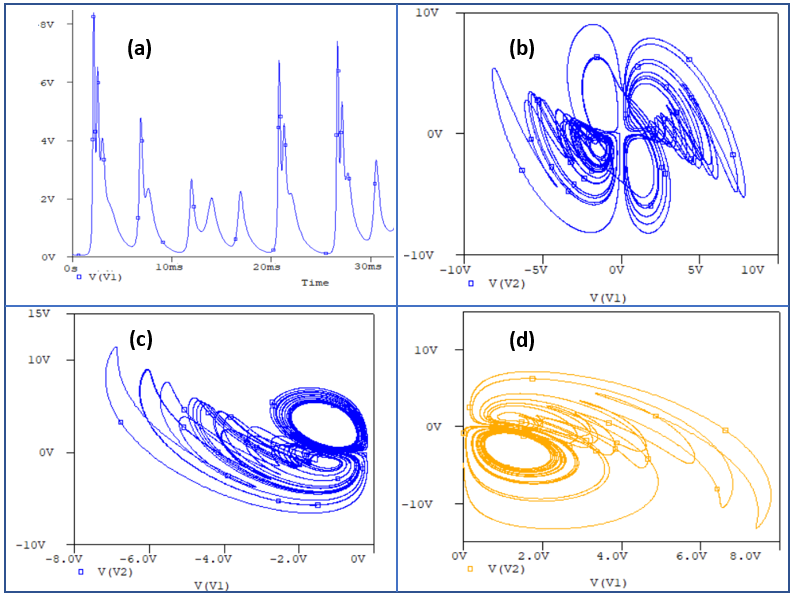}
	\caption{\textbf{(a)} Time evolution of circuit signal $v_1$, for $R_8 = 83k\Omega$ and initial voltages from the capacitors are $(1/3, -1,0)\enspace volts$. Circuit signal projected on the $v_1$-$v_2$ plane: \textbf{(b)} when $R_8=400k\Omega$ and initial voltages from the capacitors are $(-1/3,-1,0)\enspace volts$, \textbf{(c)} when $R_8=83k\Omega$ and initial voltages from the capacitors are $(-1/3,-1,0)\enspace volts$ and \textbf{(d)} when $R_8=83k\Omega$ and initial voltages from the capacitors are $(1/3, -1,0)\enspace volts$.}
	\label{psp1}
\end{figure}

In this section we show the viability of the system to engineering applications by its electronic circuit realization, cf.~\cite{pehlivan2019multiscroll,kapitaniak2018new,azar2017novel,wang2017new,nwachioma2017realization}. Owing to the physical limitations of electronic components and voltage sources, the following transformation is needed: 
\begin{equation}
w_{x} = {1\over s_1}x,\enspace w_{y} = {1\over s_2}y,\enspace w_{z} = {1\over s_3}z,
\end{equation}
where $s_1$, $s_2$ and $s_3$ are positive real constants. Choosing $s_1 = 3$, and $s_2=s_3=1$, we have 
\begin{equation}
\begin{split}
&\dot{w}_{x} = -a_1w_{x}+a_2w_{x}w_{z}+{a_3\over3}w_{y}w_{z}\\
&\dot{w}_{y} = a_4w_{y}-3a_5w_{x}w_{z}\\
&\dot{w}_{z} = -a_6w_{z}+3a_7w_{x}w_{y}+a_8w_{z}^2\label{sys2}.
\end{split}
\end{equation}	
\Cref{ecct} is a realization of system \eqref{sys2} using analog electronic components. Applying Kirchhoff's law to the circuit, we obtain	
\begin{equation}
\begin{split}
&{dv_1\over dt} = -{1\over R_1C_1}v_1+{1\over 10R_2C_1V}v_1v_3+{1\over 10R_3C_1V}v_2v_3\\
&{dv_2\over dt} = {1\over R_4C_2}{R_{10}\over R_9}v_2-{1\over 10R_5C_2V}v_1v_3\\
&{dv_3\over dt} = -{1\over R_6C_3}v_3+{1\over 10R_7C_3V}v_1v_2+{1\over 10R_8C_3V}v_3^2,\label{ccteqn}
\end{split}
\end{equation}
where $v_1$, $v_2$ and $v_3$ are the outputs of the op-amps \textit{J1A, J1B} and \textit{J1C}, respectively and they correspond to $w_{x}$, $w_{y}$ and $w_{z}$, respectively. V denotes volts. If we take $C=C_1=C_2=C_3$, we can write \eqref{ccteqn} as
\begin{equation}
\begin{split}
&{dv_1\over dt} = {1\over RC}\bigg[-{R\over R_1}v_1+{R\over 10R_2V}v_1v_3+{R\over 10R_3V}v_2v_3\bigg]\\
&{dv_2\over dt} = {1\over RC}\bigg[{R\over R_4}{R_{10}\over R_9}v_2-{R\over 10R_5V}v_1v_3\bigg]\\
&{dv_3\over dt} = {1\over RC}\bigg[-{R\over R_6}v_3+{R\over 10R_7V}v_1v_2+{R\over10R_8V}v_3^2\bigg],\label{ccteqn2}
\end{split}
\end{equation}
where $R$ is a value of resistance to be determined. To aid visualization on the oscilloscope, we can adjust the time scale as
\begin{equation}
\tau = \kappa t\label{timescale},
\end{equation}
where
\begin{equation}
\kappa:= {1\over RC}\label{timescaleconst}.
\end{equation}
With $W_i$ corresponding to $v_i$, $(i=1,2,3)$, and using \eqref{timescale} in \eqref{ccteqn2}, we can obtain the following dimensionless system of equations:
\begin{equation}
\begin{split}
&{dW_1\over d\tau} = -{R\over R_1}W_1+{R\over 10R_2}W_1W_3+{R\over 10R_3}W_2W_3\\
&{dW_2\over d\tau} = {R\over R_4}{R_{10}\over R_9}W_2-{R\over 10R_5}W_1W_3\\
&{dW_3\over d\tau} = -{R\over R_6}W_3+{R\over 10R_7}W_1W_2+{R\over 10R_8}W_3^2.\label{ccteqndimensionless}
\end{split}
\end{equation}
Now, choosing the time scale $\kappa=1000$ and $C=1nF$, then $R = 1M\Omega$ according to \eqref{timescaleconst}. Also, we can choose $R_9=R_{10} = 100k\Omega$. Therefore, comparing \cref{sys2,ccteqndimensionless}, for
\begin{enumerate}
	\item $a_j\in\mathbb{A}\enspace\forall\enspace j$, we obtain the following circuital resistances: $R_1 = 1M\Omega$, $R_2 = 100k\Omega$, $R_3=130k\Omega$, $R_4=500k\Omega$, $R_5=33k\Omega$, $R_6=166k\Omega$, $R_7=33k\Omega$, $R_8 = 400k\Omega$\label{RV1}.	
	\item
	$a_8\in\mathbb{A}$ replaced as $a_8=1.2$, we obtain the resistances: $R_8 = 83k\Omega$ and other resistances as in \cref{RV1}\label{RV2}. 
\end{enumerate}
The circuit simulation can be performed by \textit{PSpice-A/D} and \cref{psp1} shows selected signals from the simulation. \cref{psp1}(a) is the time-series signal $v_1$. The circuit signals projected on the $v_1$-$v_2$ plane are shown in: \cref{psp1}(b) corresponding to the situation where $a_8<0$ as described in \cref{secmath}, in \cref{psp1}(c) and \cref{psp1}(d) showing that bi-stability persists in the circuit realization. Hence, the attractor followed by the electronic circuit system is dependent on initial voltages on the capacitor C1, given the appropriate resistances of $R_8$. Practically, this can be achieved by reversing a polarized capacitor corresponding to the $w_x$ sub-circuit. The circuit realization can faithfully represent the system's features similar to results from numerical solutions.

\begin{figure}[htbp]
	\includegraphics[width=\linewidth]{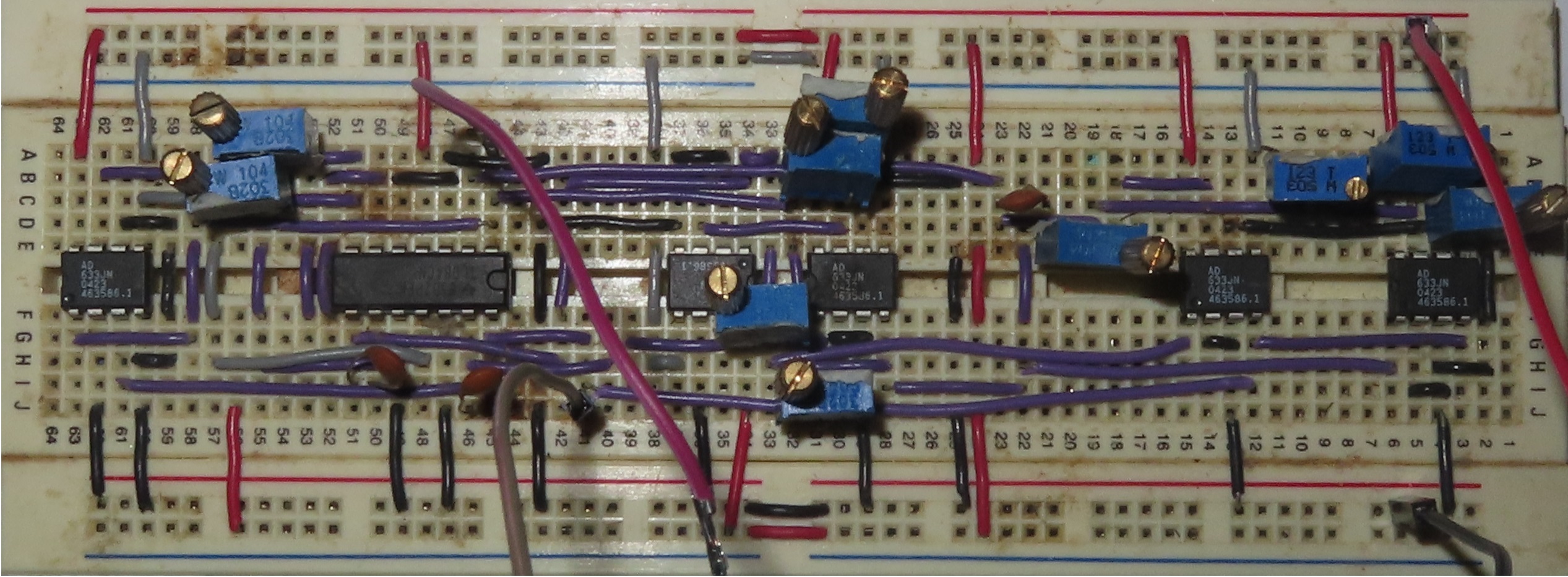}
	\caption{Analog electronic hardware implementation of the new chaotic system.}
	\label{hwimp}
\end{figure}

\begin{figure}[htbp]
	\includegraphics[width=\linewidth]{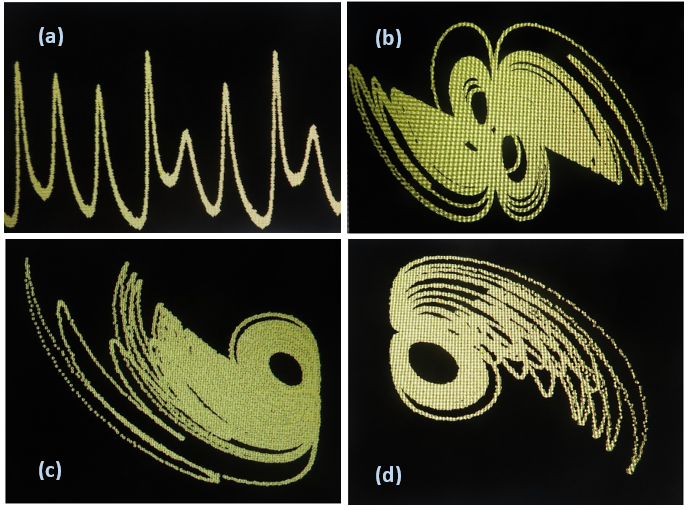}
	\caption{Circuit responses -- \textbf{(a)} Time evolution of circuit signal $v_1$, for $R_8 = 83k\Omega$. Channels 1 and 2 of the oscilloscope connected to $v_1$ and $v_2$ components of the circuit signal, respectively: \textbf{(b)} when $R_8=400k\Omega$; \textbf{(c)} when $R_8=83k\Omega$ and the case where $v_1$ component only reaches negative voltages; \textbf{(d)} when $R_8=83k\Omega$ and the case where $v_1$ component only reaches positive voltages.}
	\label{hwtesis}
\end{figure}

The system can be implemented using analog electronic components. \cref{hwimp} shows the circuit implementation, cf.~\cite{nwachioma2019new}. In particular, analog multipliers $AD633JN$ were used to realize the nonlinear signal interactions, $TL084CN$ containing four op-amps was used for signal inversion and amplification, multi-turn trimpots were used to provide the appropriate resistances. The capacitances were each $1nF$ and a symmetrical DC voltage source of $15~volts$ was used to power the circuit. Results captured on the oscilloscope are shown in \cref{hwtesis}.
\section{Chaos-based autonomous navigation of mobile robot}
An application of the new chaotic system in the control of an autonomous mobile robot is presented. The control signals are states of the new chaotic system. Mobile wheel robots for firefighting \cite{morales2011comparative}, patrol \cite{martins2007patrol} and search activities can be propelled by chaotic dynamics to realize autonomous navigation \cite{moysis2020chaotic,Volos2012a}. Due to the innate properties of chaos including sensitivity on initial condition and topological transitivity and dense orbits, dynamics of the mobile robot can become unpredictable and guaranteed to cover the entire workspace \cite{nakamura2001chaotic,volos2012c}. Furthermore, other motivating factors for using chaotic control signals for autonomous navigation of a wheeled robot are that the method is sparsely algorithmic and requires minimal programming efforts, and no huge strain on hardware resources. It can be prototyped on one of the most basic embedded platforms like the Arduino Uno device \cite{volos2013experimental}. Consequently, the method is promising in terms of energy efficiency. 

\begin{figure}[htbp]
	\includegraphics[width=0.45\textwidth]{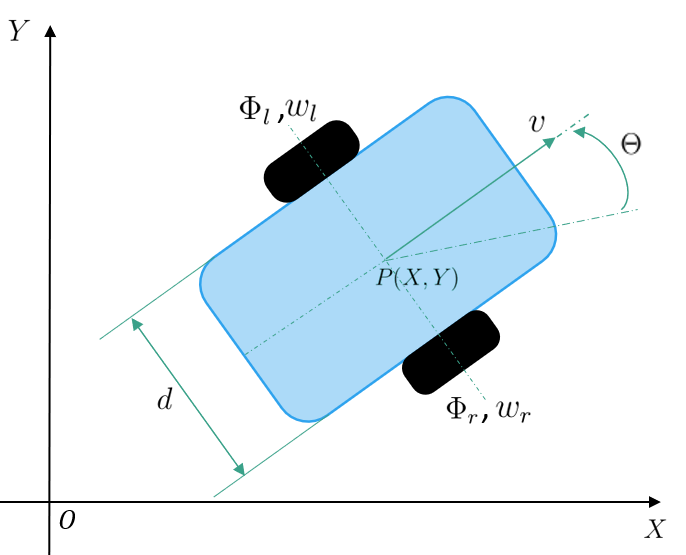}		
	\caption{Kinematics of mobile robot with differential wheels.}
	\label{mobilerobot}
\end{figure}

\Cref{mobilerobot} describes the kinematics of a differential drive mobile robot in which the wheels are powered independently by chaotic signals. Defining $(X, Y,\Theta)$ as the generalized coordinate, $P(X, Y)$ is the origin on the $X$-$Y$ plane, and $\Theta$ is the turning angle of the robot. $\Phi_l$ and $\Phi_r$ are angles of rotation of the left and right wheels, respectively, where $w_l$ and $w_r$ are the corresponding angular velocities of the wheels. $v$ is the forward velocity of the robot and $d$ is the distance between the wheels. By the principle of kinematics, we can show how the separate chaotic signals driving the wheels map to the robot's velocity.
Although the chaotic system which provides the control inputs remains uncoupled to the robot dynamics, the robot's initial position on the workspace significantly impacts the paths followed by the robot in what might be dubbed sensitivity on initial position. The robot of interest in the current study is a differential drive mobile robot with nonholonomic wheel constraints. While the velocities of the wheels can be independently controlled, they are not permitted to skid sideways. This implies that the robot is subject to the Pfaffian velocity constraint:
\begin{equation}
A(\bm{q})\dot{\bm{q}}=0,
\end{equation}
where $A(\bm{q})$ is a $k\times n$ constraint matrix and $\bm{q}\in\mathbb{R}^n$ is the configuration space of the system.
\subsection*{Assumptions:}
\begin{enumerate}
	\item The robot has a single rigid-body chassis.
	\item The robot rolls without sideways skidding on a horizontal, flat, hard surface.
	\item Frictions on the plane and relevant joints are optimal for motion without significantly impeding it. No air resistance or other forms of friction on the robot are present. 
\end{enumerate}
The robot's configuration space $\bm{q}$ is defined by
\begin{equation}
\bm{q} := (X,Y,\Theta)
\end{equation}
and the generalized velocity is therefore,
\begin{equation}
\dot{\bm{q}} = \left[\begin{array}{c}
\dot{X}\\ \dot{Y}\\ \dot{\Theta}
\end{array}\right].
\end{equation}
For a given orientation of the robot, it can be seen from \cref{mobilerobot} that the robot's velocity components are
\begin{equation}
\dot{X} = v\cos(\Theta)\label{vel1},
\end{equation} 
\begin{equation}
\dot{Y} = v\sin(\Theta)\label{vel2}
\end{equation}
and 
\begin{equation}
\dot{\Theta} = 0.
\end{equation}
From \eqref{vel2}, $v = \dot{Y}/\sin(\Theta)$. Substituting in \eqref{vel1}, we have the Pfaffian form:
\begin{equation}
\dot{X}\sin(\Theta)-\dot{Y}\cos(\Theta)+\dot{\Theta}\times0 = 0\label{velconstraint},
\end{equation}
where the constraint matrix is
\begin{equation}
A(\bm{q}) = \left[\begin{array}{ccc}
\sin(\Theta)&-\cos(\Theta)&0
\end{array}\right].
\end{equation}
Since the velocity constraint \eqref{velconstraint} cannot be integrated to give an equivalent configuration constraint, the constraint is precisely nonholonomic. Hence, the assumption of no sideways skidding.

\begin{figure}[htbp]
	\includegraphics[width=0.45\textwidth]{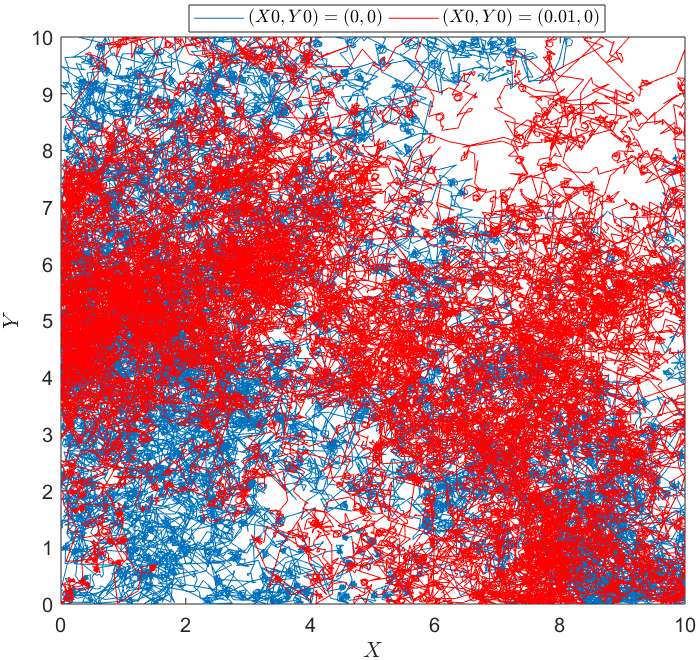}
	\caption{Trajectories of the mobile robot independently covering the $X$-$Y$ plane for cases where initial positions on the plane differ slightly.}
	\label{robotpathA}
\end{figure}

Next, to relate the forward-backward speed to the angular speed of the wheels, we consider two models, viz. the differential drive robot model and the unicycle model. In the former, the velocities are described by the system of equations:
\begin{equation}
\begin{split}
&\dot{X} = \mathcal{R}{w_r+w_l\over2}\cos(\Theta)\\
&\dot{Y} = \mathcal{R}{w_r+w_l\over2}\sin(\Theta)\\
&\dot{\Theta} = \mathcal{R}{w_r-w_l\over d},\label{ddrmodel}
\end{split}
\end{equation}
where $\mathcal{R}$ is the radius of each wheel, $w_{r}$ and $w_{l}$ are the independent angular velocities of the right and left wheels, given by
\begin{equation}
\begin{split}
w_{r} = {d\Phi_{r}\over dt}
\end{split}
\end{equation}
and 
\begin{equation}
\begin{split}
w_{l} = {d\Phi_{l}\over dt},
\end{split}
\end{equation}
respectively. Given the above assumptions, $w_{r}$ and $w_{l}$ are proportional to the applied voltages provided the voltages do not exceed the maximum that the actuators can handle. In the unicycle model, the dynamics is described by
\begin{equation}
\begin{split}
&\dot{X} = v\cos(\Theta)\\
&\dot{Y} = v\sin(\Theta)\\
&\dot{\Theta} = \mu,\label{unicycle}
\end{split}
\end{equation}
where $\mu$ is a rate of change of the robot's orientation and $v$ is the forward-backward velocity. These parameters $v$ and $\mu$ are the input signals.  From the point of view of Kinematics, both models are functionally equivalent. Thus, we can write 
\begin{equation}
v = \mathcal{R}{w_l+w_r\over2}
\end{equation}
and 
\begin{equation}
\mu = \mathcal{R}{w_l-w_r\over d}.
\end{equation}
The speed of each wheel can be independently varied depending on the applied voltage. With constant linear voltages to the left and right wheel actuators, the robot will follow a specific course on the plane, and the input velocities $v$ and $\mu$ are unchanged for the duration of a given pair of input voltages. This cannot drive the mobile robot to cover a reasonable portion of a workspace independently and randomly. To obtain a more interesting random-like motion of the robot, which could satisfactorily cover the workspace, we can modify the input velocities (or voltages) as
\begin{equation}
v := {|x+y|\over2}\label{velocity}
\end{equation}
and 
\begin{equation}
\mu := {x-y\over d}\label{angularVel},
\end{equation}
where $x$ and $y$ are variables of system \eqref{sys} and we have assumed that $\mathcal{R}=1$. Therefore, we can write a $6$-dimensional system of equations which describes the motion of the robot under the influence of the chaotic dynamics as
\begin{equation}
\begin{split}
&\dot{x} = -a_1x+a_2xz+a_3yz\\
&\dot{y} = a_4y-a_5xz\\
&\dot{z} = -a_6z+a_7xy+a_8z^2\\\label{sysrobot}
&\dot{X} = v\cos(\Theta)\\
&\dot{Y} = v\sin(\Theta)\\
&\dot{\Theta} = \mu.
\end{split}
\end{equation}
Integrating $\mu$ gives $\Theta$, which is used to determine the turning angle of the robot on the $X$-$Y$ plane. Also, the speed $v$ of the robot's wheels are determined from the chaotic signals, and $v$ has to be kept within reasonable velocities as determined by the corresponding chaotic voltages of $x$ and $y$, which are bounded. For the current simulation, a modular division can be used to achieve that. Particularly, we have made the transformation
\begin{equation}
|x+y|\rightarrow{mod \bigg(|x+y|, |x|_{max}\bigg)},
\end{equation}
where $mod(\cdot,\cdot)$ denotes the remainder when the first input entry is divided by the second entry. Hence, voltages (or proportionally speeds) to the robot's wheels are upper-bounded by $|x|_{max}$. Notice that the chaotic system is independent of the robot's motion and remains unaffected for all time $t\ge0$. Hence, the chaotic system is like the master system whereas the robot whose dynamics is being driven by the chaotic system is like the slave system in a master-slave control topology, though the robot dynamics does not necessarily track the dynamics of the chaotic system. Assuming that the distance between wheels of the mobile robot is $d=0.08$ unit and that the initial condition of system \eqref{sysrobot} is $(0.1,-0.1,0,0,0,0)$, \cref{robotpathA} shows dynamics on the $X$-$Y$ plane of the mobile robot on a workspace of dimensions $10\times10$ squared-length in normalized unit cells. The path followed by the robot can indicate sensitivity to initial position as slightly changing the previous initial condition only on the $X$-$Y$ plane, a clearly different path from the previous would be plotted. Without an operator's input, the autonomous mobile robot can cover up to $82\enspace\%$ of the workspace after a small simulation time. Due to topological mixing and dense periodic orbit of system \eqref{sys}, furthering the simulation, the robot can scan the entire workspace subject to \cref{remNavRule}.
\begin{remark}
	If obstacles are present on the workspace and an avoidance mechanism is realized (or even if not and the robot bumps into the obstacle), this will influence the overall trajectory of the autonomous mobile robot since it acquires a property, which might be called sensitivity on initial position. This is quite analogous to a common robot planning a new path when it encounters an obstacle. 
\end{remark}
\begin{remark}
	\label{remNavRule}
	A navigation rule can be defined to ensure the robot stays within the workspace of interest. For example, precedence could be given to a sensory response corresponding to \textit{no-motion} when a chaotic command signal would navigate the robot out of the workspace, cf.~\cite{moysis2020analysis}.
\end{remark}
\section{Conclusion}
A new chaotic system is proposed and its dynamical properties such as Lyapunov spectrum and dimension, multistability, a basin of attraction including the class and size of it, and bifurcation analysis are presented. It is shown that the system has bi-stable attracting sets for a specific range of the control parameter. Moreover, a circuit model of the system is realized and hardware implementation is made using off-the-shelf analog electronic components. Furthermore, it is demonstrated that the system is a good candidate for independent navigation of a differential drive mobile robot, which could be used for search, firefighting, or patrol activities. Besides, due to the robustness of the system in terms of implementation and dynamical properties, future research direction could entail using the system to innovate a computer based on the principles of chaos, cf.~\cite{pappu2020simultaneous}.
\section*{Data availability}
The datasets generated and/or analyzed during the current study are available from the first author on request.	
\section*{Conflicts of interests}
The authors declare that there is no conflict of interest regarding the publication of this paper.	
\section*{Acknowledgments}
Authors recognize the invaluable help from Dr. J. C. Sprott; particularly, Figs. 4 and 5 are courtesy of him. First author thanks CONACYT for supporting his doctoral research through the award of a scholarship. The second author recognizes the support of EDI-IPN and SNI-CONACYT. This work was financed by SIP Instituto Politécnico Nacional [grant numbers 20190052 and 20200083]


\bibliographystyle{elsarticle-num} 
\bibliography{elseviercsfbib}





\end{document}